\pdfoutput=1
\documentclass[11pt]{article}

\usepackage[final]{acl}

\usepackage{times}
\usepackage{latexsym}

\usepackage[T1]{fontenc}
\usepackage[utf8]{inputenc}

\usepackage{microtype}

\usepackage{inconsolata}

\usepackage{graphicx}

\usepackage{adjustbox}
\usepackage{booktabs}
\usepackage{graphicx}
\usepackage{amssymb}
\usepackage{amsmath}
\usepackage{algorithm}
\usepackage{algorithmicx}
\usepackage{algpseudocode}

\usepackage{paralist}

\title{RedundancyLens: Revealing and Exploiting Visual Token Processing Redundancy for Efficient Decoder-Only MLLMs}

\author{
Hongliang Li$^{1}$, Jiaxin Zhang$^{1}$, Wenhui Liao$^{1}$, \\
\textbf{Dezhi Peng}$^{1}$\textbf{,} \textbf{Kai Ding}$^{2,3}$\textbf{,} \textbf{Lianwen Jin}\footnotemark[1]\hspace{1pt}$^{1,3}$ \\
$^{1}$South China University of Technology \quad $^{2}$Intsig Information Co., Ltd.\\
$^{3}$INTSIG-SCUT Joint Lab on Document Analysis and Recognition\\
{\tt\small eehongliangli@mail.scut.edu.cn},
{\tt\small eejxzhang@gmail.com},
{\tt\small eelwh@mail.scut.edu.cn} \\
{\tt\small pengdzscut@foxmail.com},
{\tt\small danny\_ding@intsig.net},
{\tt\small eelwjin@scut.edu.cn}
}

\begin{document}
\maketitle
\begin{abstract}

\renewcommand{\thefootnote}{\fnsymbol{footnote}}
\footnotetext[1]{Corresponding author.}
\renewcommand{\thefootnote}{\arabic{footnote}}

Current Multimodal Large Language Model (MLLM) architectures face a critical tradeoff between performance and efficiency: decoder-only architectures achieve higher performance but lower efficiency, while cross-attention-based architectures offer greater efficiency but lower performance. The key distinction lies in how visual tokens are processed. Decoder-only architectures apply self-attention and FFN operations on visual tokens, while cross-attention architectures skip these computations.
To investigate whether redundancy exists in this computationally expensive process, we propose a training-free framework for analyzing trained MLLMs. It consists of Probe-Activated Dynamic FFN and Hollow Attention, which enable adjustable reductions in computations for visual tokens, as well as a Layer Ranking Algorithm that prioritizes layers for these reductions.
Extensive experiments demonstrate substantial, structured, and clustered redundancy unique to decoder-only MLLMs, offering valuable insights for future MLLM architecture design. Furthermore, by leveraging our reduction framework as a training-free inference acceleration approach, we achieve performance comparable to or better than state-of-the-art methods while remaining compatible with them.
% for further acceleration
% Code will be publicly available.
Code will be publicly available at \url{https://github.com/L-Hugh/RedundancyLens}.
\end{abstract}

\section{Introduction}
\label{sec:intro}

Large Language Models (LLMs)~\cite{NEURIPS2020_1457c0d6, OPT, LLAMA, GPT4} have seen rapid advancement in recent years, attracting attention for their strong capabilities in language comprehension and reasoning. In computer vision, researchers extend LLMs with visual abilities aimed at developing Multimodal Large Language Models (MLLMs)~\cite{BLIP2, FLAMINGO, LLAVA, MINIGPT4}. These models hold significant potential for multimodal task solving and have become a prominent focus of current research. A key challenge in this area is designing effective architectures to integrate visual signals into LLMs.

\begin{figure}[t]
  \centering
   \includegraphics[width=1.0\linewidth]{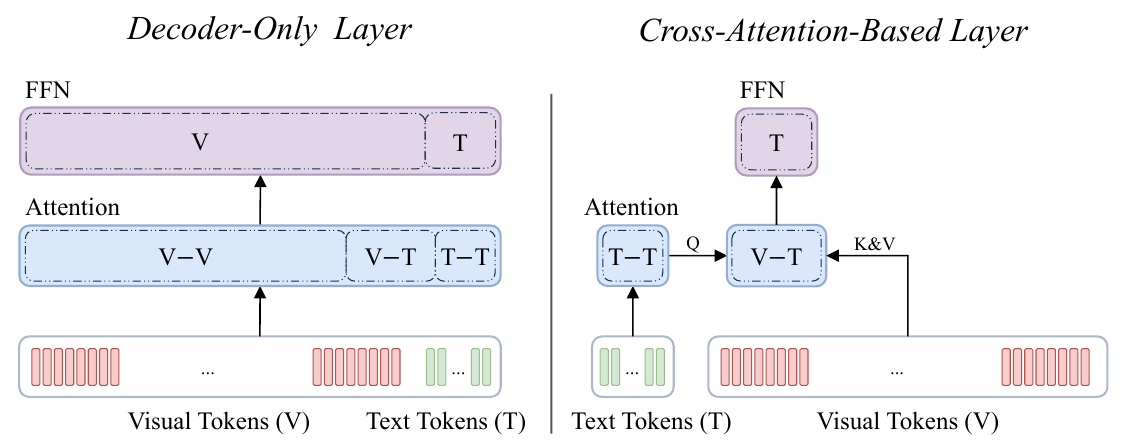}
   \caption{Comparison between decoder-only and cross-attention-based architectures from a unified perspective. Self-attention and FFN operations for visual tokens dominate the computation of decoder-only layers.}
   \label{fig:motivation}
\end{figure}

Current MLLMs are commonly built using either decoder-only (e.g., LLaVA~\cite{LLAVA}) or cross-attention-based architectures (e.g., Flamingo~\cite{FLAMINGO}). 
In the early development of MLLMs, the simplicity and effectiveness of the decoder-only architecture leads to its widespread adoption~\cite{DEEPSEEKVL, LLaVAUHD, VARY, INTERNVL}. To capture finer-grained visual details, decoder-only MLLMs progressively increase input image resolutions, resulting in significant performance gains~\cite{MONKYE, UREADER, INTERNLM4KHD}. However, this also leads to longer visual token sequences, significantly degrading the model's efficiency. Consequently, cross-attention-based architectures are attracting increasing interest due to their greater efficiency in handling long visual token sequences~\cite{LLAMA3, EVLM}. Nonetheless, recent work~\cite{NVLM} demonstrates that, decoder-only MLLMs tend to achieve significantly better overall performance. Developing an MLLM architecture that achieves both high performance and efficiency remains an important area for further research.

To advance this research, this paper investigates MLLM architectures by evaluating existing designs and analyzing their redundancy patterns. We begin by comparing these two common architectures from a unified perspective.
As shown in Figure~\ref{fig:motivation}, the fundamental difference between them lies in the processing
of visual tokens: in decoder-only architectures, visual tokens undergo self-attention and FFN operations, whereas cross-attention-based architectures omit these operations.
Since visual tokens often outnumber text tokens significantly~\cite{INTERNLM4KHD}, the processing of visual tokens consumes the majority of computational resources (roughly estimated as the ratio of visual tokens to total tokens, typically exceeding 90\%). 
Investigating whether redundancy exists in this computationally expensive process is valuable.
Considering the computational cost can be expressed as the number of layers multiplied by the cost of performing self-attention and FFN operations on visual tokens at each layer, the question arises: \textit{are full self-attention and FFN operations for visual tokens required at every layer?}

Given the significant training costs of state-of-the-art MLLMs, we propose a training-free framework to investigate this question by analyzing trained decoder-only MLLMs. Specifically, we apply computational reductions to visual token processing in a subset of layers and evaluate their impact on model performance. By gradually increasing the number of layers where these reductions are applied, from a single layer to all layers, we can obtain a performance variation curve that reflects the degree of redundancy in the self-attention and FFN operations across layers.

To achieve this, the proposed framework consists of two components:
(1) Probe-Activated Dynamic FFN and Hollow Attention, which replace the original FFN and attention modules, enabling adjustable reductions in computations for visual tokens. Specifically, Probe-Activated Dynamic FFN dynamically selects a subset of FFN parameters to process visual tokens. The Probe-Activated strategy is proposed to enable this selection in a training-free manner. Hollow Attention limits global attention among visual tokens to local attention while preserving attention between visual and text tokens.
(2) Layer Ranking Algorithm, which assigns a rank to each layer. When selecting a subset of layers for computational reductions during the traversal process, those with the highest ranks are prioritized.

We conduct extensive experiments on state-of-the-art MLLMs, including InternVL2-8B~\cite{INTERNVL15}, Qwen2-VL-7B~\cite{QWEN2VL}, MiniCPM-V 2.6~\cite{MINICPMV}, and LLaVA-OneVision~\cite{LLAVAONEVISION}. Our experiments are divided into two parts.

For the first part, the results show that applying the proposed reductions to approximately half of the layers preserves or even improves model performance. Notably, further applying these reductions to text tokens leads to a sharp decline in model performance. These findings reveal that decoder-only MLLMs exhibit substantial redundancy in the processing of visual tokens within certain layers. This structured and clustered redundancy can be effectively leveraged, providing valuable insights for future architecture design.

For the second part, leveraging our reduction framework as a training-free inference acceleration
approach, we achieve performance comparable to or better than current state-of-the-art methods~\cite{FASTV, VTW}. Furthermore, existing approaches accelerate models by reducing the number of visual tokens, while our approach reduces the computational cost per visual token. Since these two methods are orthogonal, they can be combined for further acceleration.

In conclusion, our contributions are three-fold:
\begin{itemize}
\item We propose a framework to investigate redundancy in visual token processing through the analysis of trained decoder-only MLLMs.
\item We demonstrate substantial, structured, and clustered redundancy unique to decoder-only MLLMs, offering valuable insights for future MLLM architecture design. 
\item We introduce a training-free MLLM acceleration method that takes a distinct and orthogonal perspective from current state-of-the-art methods, achieving comparable or better results while remaining compatible with them.
\end{itemize}

\section{Related Work}
\label{sec:related}

%-------------------------------------------------------------------------
\subsection{MLLM Architectures}

The decoder-only architecture is one of the most widely adopted designs for MLLMs~\cite{BLIP2, LLAVA, MINIGPT4}, favored for its simplicity and efficiency. In this architecture, image tokens are concatenated with text token sequences and processed uniformly alongside text tokens by the LLM. A projector module maps the features extracted by the image encoder into the input image tokens for the LLM, implemented using either a multilayer perceptron~\cite{LLAVA, COGVLM, DEEPSEEKVL} or cross-attention mechanisms~\cite{BLIP2, QWENVL, MPLUGOWL}. To improve fine-grained visual perception by capturing more detailed visual features, models like UReader~\cite{UREADER} and Monkey~\cite{MONKYE} divide high-resolution images into multiple sub-images and concatenate their tokens for input into the LLM. Extending this idea, InternLM-XComposer2-4KHD~\cite{INTERNLM4KHD} enhances the model's resolution capabilities to 4K HD and beyond, demonstrating consistent performance improvements. These advances have significantly accelerated the development of MLLMs~\cite{INTERNVL15, MINICPMV, COGVLM2, QWEN2VL, LLAVAONEVISION}, allowing open-source models to match or even surpass commercial multimodal models. However, increasing image resolution and multi-image input scenarios lead to longer input sequences, which significantly increase inference times and limit practical applications.

The cross-attention-based architecture offers greater efficiency in handling long visual token sequences, gaining increasing attention as an alternative to the decoder-only architecture. These architectures introduce additional cross-attention layers within the LLM to integrate visual information by applying cross-attention to visual tokens, thereby eliminating the need for the entire LLM to process them. Flamingo~\cite{FLAMINGO} is a prominent early work in this area, using a perceiver resampler to downsample the vision encoder’s features before feeding them into the LLM via gated cross-attention layers. Llama 3-V~\cite{LLAMA3} adopts a similar structure but removes the perceiver module. EVLM~\cite{EVLM} utilizes hierarchical ViT features and a mixture of experts to enhance performance. mPLUG-Owl3~\cite{MPLUGOWL3} incorporates cross-attention mechanisms in parallel with self-attention layers instead of adding additional cross-attention layers. EE-MLLM~\cite{EEMLLM} modifies the original self-attention mechanism into a composite attention mechanism. Meanwhile, NVLM~\cite{NVLM} introduces a hybrid architecture that uses the LLM’s self-attention layers to process thumbnail image tokens while employing cross-attention to capture finer image details.

To provide a fair comparison of the two architectures, recent work~\cite{NVLM} trained both a decoder-only MLLM (NVLM-D) and a cross-attention-based MLLM (NVLM-X) under the same conditions. The results show that NVLM-X provides superior computational efficiency for high-resolution images, whereas NVLM-D delivers better overall performance. 
This comparison provides valuable insights for future research; however, further investigation at a more granular level would be beneficial.

\subsection{Visual Token Compression in MLLMs}

Compressing visual sequence length is an effective and common method for accelerating MLLMs~\cite{liu2024visualanchorsstronginformation, zhang2024dockylinlargemultimodalmodel, xing2024pyramiddrop, huang2024mini, he2024zipvl}. Common techniques include using a group of learnable query tokens to extract information via cross-attention~\cite{InstructBLIP, BLIP2, FLAMINGO}, directly concatenating adjacent tokens~\cite{MINIGPTV2, TEXTHAWK}, or downsampling through convolutional neural networks~\cite{HONEYBEE, MPLUGDOCOWL15}. Some recent approaches dynamically discard nonessential tokens during inference~\cite{PRUMERGE, FASTV, VTW}. For instance, FastV~\cite{FASTV} reduces computational costs dramatically by pruning visual tokens based on their average attention scores at a selected layer in the MLLM, without sacrificing performance.

Token compression methods accelerate MLLMs by reducing the number of visual tokens, but the remaining tokens still require substantial computation in the LLM module, similar to text tokens. In contrast, our method achieves acceleration by reducing computation per visual token. 
This means our method is orthogonal to these methods and can be combined with them for further acceleration.

\begin{figure*}
  \centering
  \includegraphics[width=0.95\linewidth]{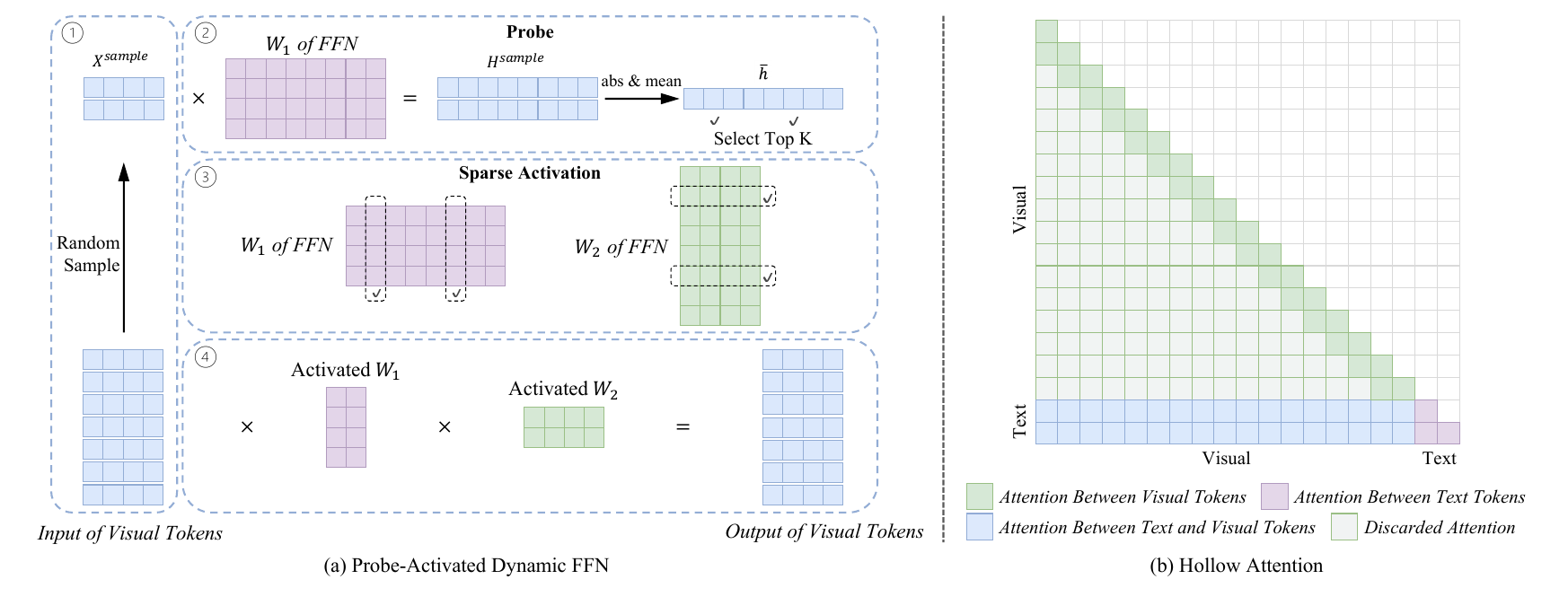}
  \caption{Illustration of the proposed computational reductions for visual tokens: (a) Probe-Activated Dynamic FFN and (b) Hollow Attention. The symbol `×' denotes matrix multiplication.}
  \label{fig:ov}
\end{figure*}

\section{Methodology}
\label{sec:metho}

The proposed framework consists of two components:
(1) Probe-Activated Dynamic FFN and Hollow Attention, which replace the original FFN and attention modules, enabling adjustable reductions in computations for visual tokens.
(2) Layer Ranking Algorithm, which assigns a rank to each layer. When selecting a subset of layers for computational reductions, those with the highest ranks are prioritized.

\subsection{Computational Reductions for Visual Tokens}

\subsubsection{Probe-Activated Dynamic FFN}

Inspired by MoE~\cite{shazeer2016outrageously}, we reduce FFN computations for visual tokens by structurally activating only a subset of FFN parameters. However, we cannot directly adopt MoE, as it requires training a router that dynamically selects which parameters to activate. To achieve this without additional training, we propose Probe-Activated Dynamic FFN.

For each forward pass, the visual input  \(X \in \mathbb{R}^{N \times d_{\text{model}}}\) consists of \(N\) visual tokens, derived from a single image or multiple cropped images, where \(d_{\text{model}}\) is the feature dimension. The vanilla FFN layer \cite{TRANSFORMER} performs the following operations to obtain the output \(Y\):
\begin{equation}
  H = \text{ReLU}(X W_1 + \mathbf{b_1}) \in \mathbb{R}^{N \times d_{\text{ff}}},
  \label{eq:ffn1}
\end{equation}
\begin{equation}
  Y = H W_2 + \mathbf{b_2} \in \mathbb{R}^{N \times d_{\text{model}}},
  \label{eq:ffn2}
\end{equation}
where \(W_1 \in \mathbb{R}^{d_{\text{model}} \times d_{\text{ff}}}\) and \(W_2 \in \mathbb{R}^{d_{\text{ff}} \times d_{\text{model}}}\) are the weight matrices.

In the proposed Probe-Activated Dynamic FFN, we first randomly sample a subset \(X^{\text{sample}} \in \mathbb{R}^{M \times d_{\text{model}}}\) from \(X\), where \(M\) (\(M \ll N\)) denotes the number of sampled tokens. This sampled subset is used to compute the hidden representation:
\begin{equation}
  H^{\text{sample}} = \text{ReLU}(X^{\text{sample}} W_1 + \mathbf{b_1}) \in \mathbb{R}^{M \times d_{\text{ff}}}.
  \label{eq:sub1}
\end{equation}
We then take the element-wise absolute value of each token's hidden representation and compute the mean across the sampled tokens:
  \begin{equation}
    \bar{\mathbf{h}} = \frac{1}{M} \sum_{i=1}^{M} \left| H^{\text{sample}}_i \right| \in \mathbb{R}^{d_{\text{ff}}}.
  \label{eq
}
\end{equation}
Next, we select the top \(K\) elements from \(\bar{\mathbf{h}}\) with the highest values. Let \(S\) represent the set of selected indices:
\begin{equation}
  S = \text{Top}_K(\bar{\mathbf{h}}).
  \label{eq:sub3}
\end{equation}
Using the selected indices \(S\), we activate a subset of the weight matrices \(W_1\) and \(W_2\) as follows:
\begin{equation}
  W_1^{\text{act}} = W_1[:, S] \in \mathbb{R}^{d_{\text{model}} \times K},
  \label{eq:sub4}
\end{equation}
\begin{equation}
  W_2^{\text{act}} = W_2[S, :] \in \mathbb{R}^{K \times d_{\text{model}}}.
  \label{eq:sub5}
\end{equation}
% The corresponding bias \(b_1\) is also selected as follows:
The corresponding bias \(\mathbf{b_1}^{\text{act}}\) is activated similarly:
\begin{equation}
  \mathbf{b_1}^{\text{act}} = \mathbf{b_1}[S] \in \mathbb{R}^{K}.
  \label{eq:sub6}
\end{equation}
Finally, the forward propagation proceeds as follows:
\begin{equation}
  H^{\text{act}} = \text{ReLU}(X W_1^{\text{act}} + \mathbf{b_1}^{\text{act}}) \in \mathbb{R}^{N \times K},
  \label{eq:fina1}
\end{equation}
\begin{equation}
  Y = H^{\text{act}} W_2^{\text{act}} + \mathbf{b_2} \in \mathbb{R}^{N \times d_{\text{model}}}.
  \label{eq:fina2}
\end{equation}

Figure~\ref{fig:ov} (a) provides a more intuitive illustration of the computation process in Probe-Activated Dynamic FFN, with activation functions and biases omitted for simplicity. It is important to note that this process applies only to visual tokens, while the FFN for text tokens remains unchanged. Some MLLMs modify the vanilla FFN, such as by adding gating mechanisms \cite{QWEN2VL, INTERNVL15}, yet our method can still be directly applied in these cases.

\subsubsection{Hollow Attention}

Inspired by sparse attention~\cite{BigBird}, we introduce a custom sparse attention pattern for MLLMs, called Hollow Attention, to reduce the attention computation for visual tokens. As illustrated in Figure~\ref{fig:ov} (b), global attention among visual tokens is replaced with local attention, while the attention between visual and text tokens, as well as within text tokens, remains unchanged.
Specifically, each visual token attends to the preceding \( R_A \) visual tokens (where \( R_A \) denotes the attention range) and all text tokens,
whereas text tokens retain the ability to attend to all tokens. Since visual tokens typically outnumber text tokens by a large margin in MLLMs, this reduction effectively eliminates the majority of the attention overhead.

\subsection{Layer Ranking Algorithm}

Given the number of layers requiring reduction, denoted by \(L_r\) (where \(0 \leq L_r \leq L\), and \(L\) is the total number of layers), the goal is to select the \(L_r\)-layer combination with the highest redundancy. To this end, we construct a compact validation set and use the performance variations of the MLLM on it to estimate redundancy. Since exhaustively evaluating all possible layer combinations for each value of \(L_r\) is computationally infeasible, we propose a search algorithm that ranks each layer, as detailed in Algorithm~\ref{alg:layer_ranking}. For a given \(L_r\),
the top-ranked \(L_r\) layers are selected for reduction.

\begin{algorithm}[t]
\caption{Layer Ranking Search}
\label{alg:layer_ranking}
\small
\begin{algorithmic}[1]
\State \textbf{Input:} Number of layers \( L \), validation set
\State \textbf{Output:} Ranked list of layer indices

\State \( RankedLayers \gets [] \) 
\State \( UnrankedLayers \gets \{ 1, 2, \dots, L \} \)

\While{\( UnrankedLayers \neq \emptyset \)}
    \State \( SelectedLayer \gets null \)
    \State \( BestPerformance \gets -\infty \)
    
    \For{each \(layer\) in \( UnrankedLayers \)}
        \State Apply reduction to \( RankedLayers \cup \{layer\} \)
        \State Evaluate the model on the validation set
        \State and store the performance metric as \( P \)
   
        \If{\( P > BestPerformance \)}
            \State \( BestPerformance \gets P \)
            \State \( SelectedLayer \gets layer \)
        \EndIf
    \EndFor

    \State \( RankedLayers.append(SelectedLayer) \)
    \State \( UnrankedLayers.remove(SelectedLayer) \)

\EndWhile

\State \textbf{Return:} \( RankedLayers \)
\end{algorithmic}
\end{algorithm}

The validation set used in the Layer Ranking Algorithm comprises multiple subsets drawn from different datasets. The overall evaluation metric is computed by summing the scores across all subsets. For each subset, we compute the difference by subtracting the evaluation metric of the original model from that of the reduced model. If this difference is negative—indicating that the reduced model performs worse—it is multiplied by a penalty coefficient \(\alpha > 1\). This penalization mechanism encourages the search process to prioritize reductions that maintain performance stability. 

For each MLLM, Algorithm~\ref{alg:layer_ranking} is applied separately for FFN and attention reductions, with the process run twice. In our experiments, we observe that the last few layers of MLLMs tend to exhibit greater redundancy, making them a priority for reduction. To reduce the number of evaluations, we limit the ranking algorithm’s search to the first \( L - L_p \) layers. The last \( L_p \) layers are ranked in descending order of their position, starting from the last layer.

\begin{figure*}
  \centering
  \includegraphics[width=1.0\linewidth]{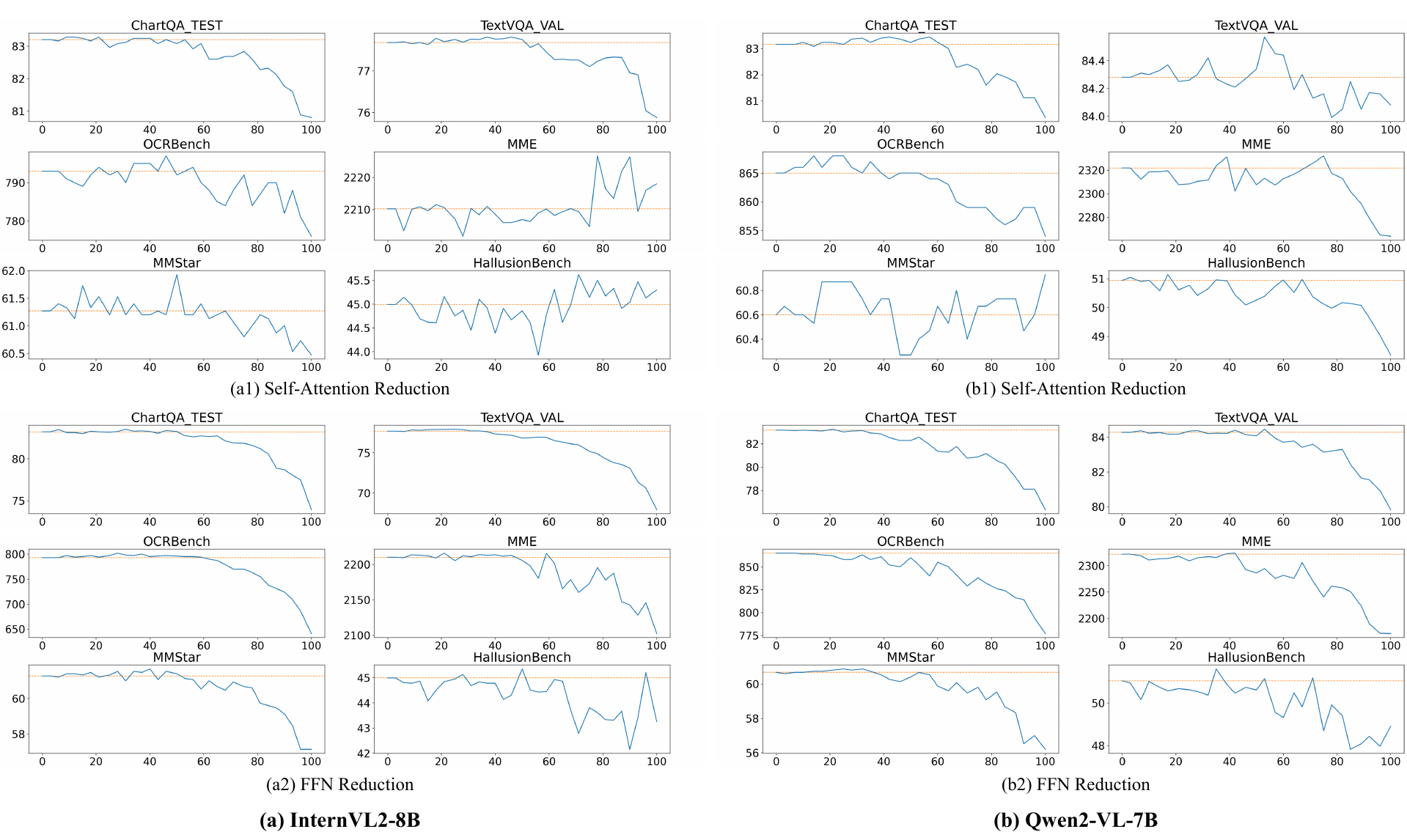}
  \caption{Impact of applying self-attention or FFN reductions across various layer proportions. The x-axis represents the percentage of layers with reductions applied, and the y-axis indicates model performance on the benchmark metric. The horizontal line shows the model's original performance (y-value at x=0).}
  \label{fig:internvl_qwen}
\end{figure*}

\section{Experiments}
\label{sec:exp}

\subsection{Datasets}

To construct the validation set for the Layer Ranking Algorithm, we randomly sample 750 instances from the full evaluation dataset collected in~\cite{OCRBench}, 1,000 instances from MMBench-DEV-EN-V11~\cite{MMBench}, and 200 instances each from the validation sets of DocVQA~\cite{DOCVQA}, InfoVQA~\cite{INFOVQA}, and ChartQA~\cite{CHARTQA}. To avoid overlap with the test set, when sampling the 750 instances from the full evaluation dataset in~\cite{OCRBench}, we exclude any samples that appear in OCRBench~\cite{OCRBench}, as well as those from TextVQA~\cite{TEXTVQA}, DocVQA~\cite{DOCVQA}, InfoVQA~\cite{INFOVQA}, and ChartQA~\cite{CHARTQA}.

For evaluation, we conduct experiments on eight widely used benchmarks: OCRBench~\cite{OCRBench}, DocVQA~\cite{DOCVQA}, InfoVQA~\cite{INFOVQA}, ChartQA~\cite{CHARTQA}, TextVQA~\cite{TEXTVQA}, MME~\cite{MME}, MMStar~\cite{MMSTAR}, and HallusionBench~\cite{HallusionBench}. In particular, following prior work~\cite{INTERNVL15, QWEN2VL}, we use the validation set of TextVQA~\cite{TEXTVQA} for evaluation.

\subsection{Evaluation Metrics for Each Benchmark}
We assess model performance using the standard metrics provided by each benchmark. OCRBench~\cite{OCRBench} uses the number of correctly generated answers as its evaluation metric. DocVQA~\cite{DOCVQA} and InfoVQA~\cite{INFOVQA} use Average Normalized Levenshtein Similarity (ANLS) and are evaluated on their respective official websites. ChartQA~\cite{CHARTQA} measures performance with relaxed accuracy, while TextVQA~\cite{TEXTVQA} relies on VQA accuracy~\cite{goyal2017making}. MME~\cite{MME} reports the sum of perception and cognition scores. MMStar~\cite{MMSTAR} evaluates models based on overall accuracy. HallusionBench~\cite{HallusionBench} reports the average of Question Pair Accuracy, Figure Accuracy, and Overall Accuracy. Evaluation for all benchmarks, except DocVQA and InfoVQA, is conducted using VLMEvalKit~\cite{VLMEVALKIT}.

\subsection{Implementation Details}
All experiments are conducted on NVIDIA A100 GPUs using VLMEvalKit~\cite{VLMEVALKIT}, a framework for evaluating MLLMs on diverse multimodal benchmarks. We evaluate state-of-the-art MLLMs, including InternVL2-8B~\cite{INTERNVL15}, Qwen2-VL-7B~\cite{QWEN2VL}, MiniCPM-V 2.6~\cite{MINICPMV}, and LLaVA-OneVision~\cite{LLAVAONEVISION}. In Hollow Attention, the attention range \( R_A \) for visual tokens is set to 256, which typically corresponds to the number of tokens in a single sub-image. In Probe-Activated Dynamic FFN, the number of randomly sampled visual tokens \( M \) is set to 10\% of the total visual tokens per sample, while the number of activated parameters \( K \) is set to 20\% of the original parameter count.
For the Layer Ranking Algorithm, the penalty coefficient \( \alpha \) is set to 2. 
During the ranking process, \( M \) is set to 100\% of the total visual tokens per sample to minimize fluctuations caused by uncertainty.

\subsection{Redundancy Analysis in FFN and Attention for Visual Tokens}

First, we independently analyze the redundancy of FFN and self-attention operations on visual tokens in existing decoder-only MLLMs. Specifically, we apply computational reductions for FFN or self-attention in a subset of layers and evaluate the MLLM's performance on multiple mainstream benchmarks. By gradually increasing the number of layers where these reductions are applied, from a single layer to all layers, we obtain performance variation curves that reflect the degree of redundancy in the self-attention and FFN operations across layers.

\begin{figure}[t]
  \centering
  \includegraphics[width=1.0\linewidth]{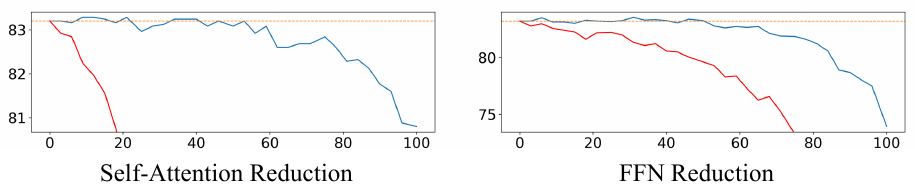}
  \caption{Performance comparison of the reductions applied to visual tokens (blue line) versus all tokens (red line), evaluated on ChartQA by InternVL2-8B.}
  \label{fig:all_token}
\end{figure}

The experimental results are shown in Figure~\ref{fig:internvl_qwen}. The results indicate that applying the proposed reductions to about half of the layers maintains the MLLM's performance at a level comparable to the original model on most benchmarks, and in some cases, even surpasses the unreduced baseline. This outcome holds true for both the InternVL2~\cite{INTERNVL15} and Qwen2-VL~\cite{QWEN2VL}, regardless of whether the reductions are applied to self-attention or FFN operations. However, when reductions are applied to more than half of the layers, the performance of the MLLMs begins to decline rapidly across most benchmarks, with FFN reductions causing a sharper drop than self-attention reductions. In addition, further applying these reductions to text tokens leads to a sharp decline in model performance, as shown in Figure~\ref{fig:all_token}. These findings reveal that decoder-only MLLMs exhibit substantial redundancy in the processing of visual tokens within certain layers.

Current state-of-the-art MLLMs are built on pre-trained LLMs and fine-tuned on vast multimodal datasets, such as Qwen2-VL~\cite{QWEN2VL}, trained on over 1.4 trillion tokens of multimodal data.
Therefore, the redundancy observed in processing visual tokens within LLMs cannot be attributed solely to insufficient training. We argue that this redundancy arises more from the inherent differences between visual and text tokens. On one hand, visual and text tokens originate from different modalities; on the other, visual tokens undergo extensive processing through an image encoder, while text tokens are processed only through linear mapping. These differences suggest that treating them equivalently within the LLM may not be the most efficient approach, especially considering the high computational demands of MLLMs in practice. By highlighting such redundancy, we hope to provide valuable insights for future architecture design.

\begin{figure}[t]
  \centering
  \includegraphics[width=1.0\linewidth]{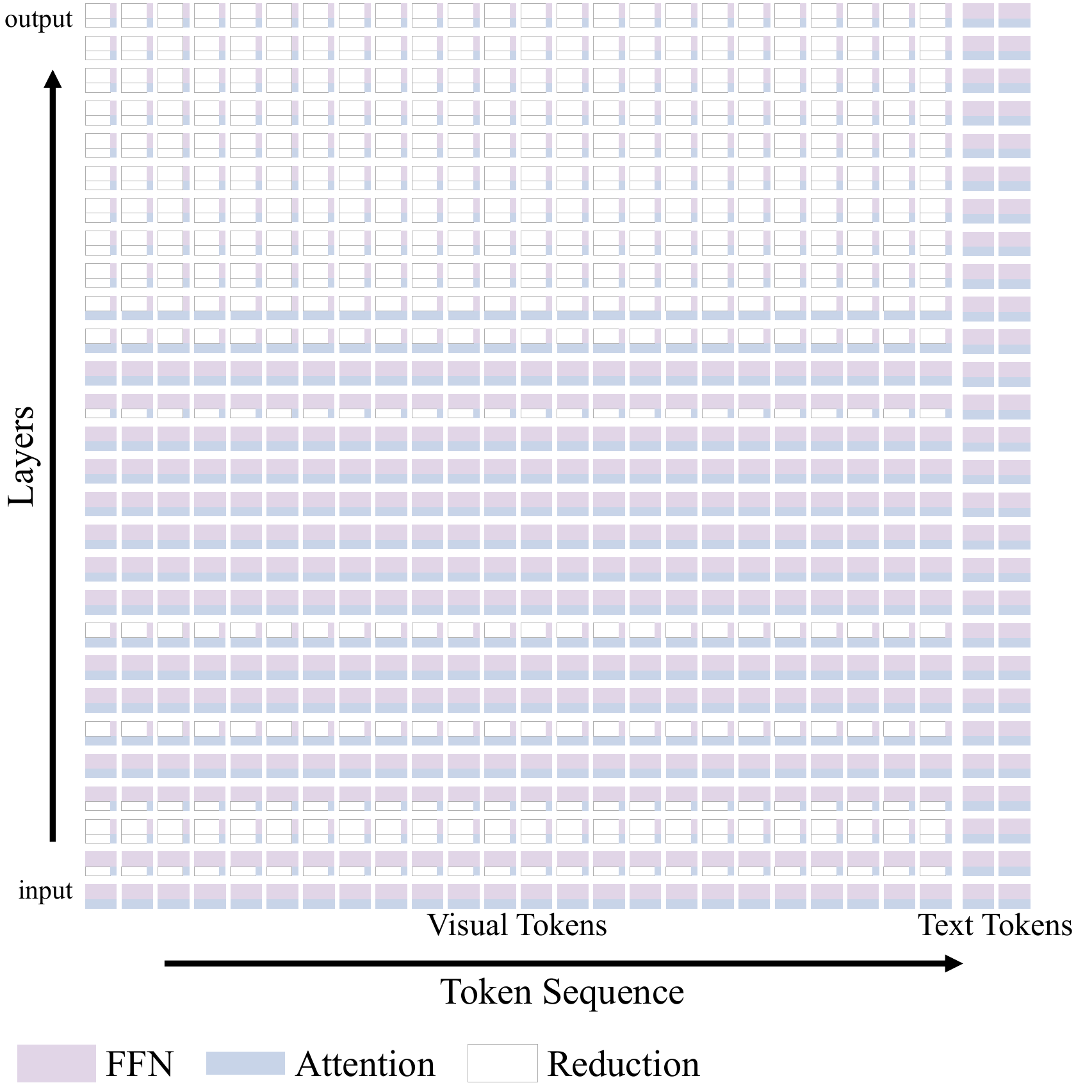}
  \caption{The reduction pattern for Qwen2-VL-7B. Notably, the attention from text tokens to all visual tokens needs to be preserved.}
  \label{fig:pattern}
\end{figure}

\begin{table*}[t]
  \centering
  \begin{adjustbox}{width=1.0\textwidth}
  \begin{tabular}{@{}lccccccccc@{}}
    \toprule
    Method & FLOPs Ratio & OCRBench & DocVQA & InfoVQA & ChartQA & TextVQA & MME & MMStar & HallusionBench\\

    \midrule
    InternVL2-8B (32 Layers) & 100\% & 793 & 91.6 & 74.4 & 83.2 & 77.7 & 2210 & 61.3 & 45.0 \\
    \midrule
    \quad + VTW~\cite{VTW} ($L$=23) & 72\% & 704 & 87.9 & 69.3 & 80.0 & 69.3 & \underline{2201} & \underline{61.2} & 44.6\\
    \quad + FastV~\cite{FASTV} ($K$=2, $R$=30\%) & 72\% & \underline{793} & \underline{90.6} & \underline{71.6} & \underline{82.9} & \textbf{77.6} & 2181 & 60.7 & \underline{45.3}\\
    \quad + \textbf{Ours} ($L_{RA}$=16, $L_{RF}$=17) & 72\% & \textbf{801} & \textbf{91.3} & \textbf{74.4} & \textbf{83.1} & \underline{77.2} & \textbf{2212} & \textbf{61.7} & \textbf{45.6}\\
    \cmidrule(lr){1-10}
    \quad + VTW~\cite{VTW} ($L$=17) & 53\% & 64 & 14.5 & 30.9 & 17.9 & 20.8 & \textbf{2200} & \underline{59.3} & 44.8\\
    \quad + FastV~\cite{FASTV} ($K$=2, $R$=50\%) & 53\% & \underline{768} & \underline{85.4} & \underline{66.1} & \underline{80.6} & \textbf{77.1} & \underline{2195} & \underline{59.3} & \underline{44.9}\\
    \quad + \textbf{Ours} + FastV ($K$=2, $R$=30\%) & 52\% & \textbf{797} & \textbf{90.3} & \textbf{71.6} & \textbf{83.0} & \textbf{77.1} & 2192 & \textbf{60.9} & \textbf{45.9}\\

    \midrule 
    Qwen2VL-7B (28 Layers) & 100\% & 865 & 94.5 & 76.6 & 83.2 & 84.3 & 2322 & 60.7 & 51.0\\
    \midrule
    \quad + VTW~\cite{VTW} ($L$=20) & 71\% & 41 & 13.7 & 31.1 & 19.4 & 15.9 & \textbf{2311} & \textbf{60.7} & \underline{50.3}\\
    \quad + FastV~\cite{FASTV} ($K$=2, $R$=30\%) & 72\% & \underline{829} & \underline{94.4} & \underline{75.1} & \underline{82.6} & \underline{84.0} & 2306 & 59.9 & 49.8 \\
    \quad + \textbf{Ours} ($L_{RA}$=13, $L_{RF}$=14)      & 71\% & \textbf{859} & \textbf{94.5} & \textbf{75.7} & \textbf{83.0} & \textbf{84.6} & \underline{2309} & \underline{60.5} & \textbf{51.1}\\
    \cmidrule(lr){1-10}
    \quad + VTW~\cite{VTW} ($L$=15) & 54\% & 36 & 8.4 & 23.8 & 16.6 & 13.7 & 2174 & 53.7 & 39.7\\
    \quad + FastV~\cite{FASTV} ($K$=2, $R$=50\%) & 53\% & \underline{766} & \underline{93.4} & \underline{71.0} & \underline{79.4} & \underline{83.6} & \underline{2309} & \underline{58.6} & \underline{49.3}\\
    \quad + \textbf{Ours} + FastV ($K$=2, $R$=30\%) & 53\% & \textbf{832} & \textbf{94.3} & \textbf{74.3} & \textbf{81.8} & \textbf{84.2} & \textbf{2310} & \textbf{59.7} & \textbf{51.1}\\

    \midrule 
    LLaVA-OneVision-7B (28 Layers) & 100\% & 623 & 87.5 & 65.0 & 80.4 & 76.0 & 2002 & 61.7 & 39.3\\
    \midrule
    \quad + VTW~\cite{VTW} ($L$=20) & 71\% & 47 & 15.2 & 30.5 & 19.8 & 15.3 & 1991 & \textbf{61.4} & \underline{40.2}\\
    \quad + FastV~\cite{FASTV} ($K$=2, $R$=30\%) & 72\% & 590 & \underline{85.4} & \underline{61.4} & \underline{77.0} & \textbf{75.1} & \underline{2007} & 60.1 & 40.1 \\
    \quad + \textbf{Ours} ($L_{RA}$=13, $L_{RF}$=14)  & 71\% & \textbf{635} & \textbf{85.6} & \textbf{63.2} & \textbf{80.2} & \underline{75.0} & \textbf{2019} & \underline{61.0} & \textbf{40.3}\\
    \cmidrule(lr){1-10}
    \quad + VTW~\cite{VTW} ($L$=15) & 54\% & 34 & 10.5 & 25.0 & 18.4 & 14.2 & 1897 & 52.5 & 29.6\\
    \quad + FastV~\cite{FASTV} ($K$=2, $R$=50\%) & 53\% & \underline{506} & \underline{78.9} & \underline{53.8} & \underline{68.0} & \underline{72.5} & \underline{1974} & \underline{57.9} & \underline{39.2}\\
    \quad + \textbf{Ours} + FastV ($K$=2, $R$=30\%) & 53\% & \textbf{597} & \textbf{84.0} & \textbf{59.3} & \textbf{76.8} & \textbf{74.4} & \textbf{2019} & \textbf{59.9} & \textbf{40.2}\\

    \midrule
    MiniCPM-V 2.6 (28 Layers) & 100\% & 846 & 90.6 & 64.6 & 80.4 & 79.2 & 2276 & 57.5 & 48.3\\
    \midrule
    \quad + VTW~\cite{VTW} ($L$=20) & 71\% & 130 & 16.8 & 31.0 & 21.3 & 20.4 & 2250 & \underline{57.3} & 34.9 \\
    \quad + FastV~\cite{FASTV} ($K$=2, $R$=30\%) & 72\% & \underline{800} & \underline{85.5} & \underline{59.3} & \underline{78.4} & \underline{79.0} & \underline{2252} & 56.3 & \underline{46.1} \\
    \quad + \textbf{Ours} ($L_{RA}$=13, $L_{RF}$=14)  & 71\% & \textbf{847} & \textbf{90.0} & \textbf{63.8} & \textbf{79.8} & \textbf{79.6} & \textbf{2274} & \textbf{57.5} & \textbf{46.7} \\
    \cmidrule(lr){1-10}
    \quad + VTW~\cite{VTW} ($L$=15) & 54\% & 113 & 12.7 & 27.6 & 18.3 & 18.1 & 2053 & 53.5 & 30.1 \\
    \quad + FastV~\cite{FASTV} ($K$=2, $R$=50\%) & 53\% & \underline{749} & \underline{72.5} & \underline{52.2} & \underline{72.9} & \underline{77.0} & \underline{2189} & \underline{54.4} & \underline{46.5} \\
    \quad + \textbf{Ours} + FastV ($K$=2, $R$=30\%) & 53\% & \textbf{805} & \textbf{84.7} & \textbf{58.9} & \textbf{78.2} & \textbf{78.8} & \textbf{2228} & \textbf{55.1} & \textbf{46.7} \\
    
    \bottomrule
  \end{tabular}
  \end{adjustbox}
  \caption{Comparison of training-free methods for accelerating MLLM inference. The \( L_{RA} \) and \( L_{RF} \) in our method represent the number of layers for attention reduction and FFN reduction, respectively. FLOPs Ratio indicates the proportion of floating-point operations retained after applying the acceleration method compared to the full model. The best results are highlighted in bold, while the second-best results are underlined.}
  \label{tab:main}
\end{table*}

\subsection{Comparison with Training-free MLLM Inference Acceleration Methods}

Building on the previous conclusions, our framework can accelerate decoder-only MLLM inference in a training-free manner, with the reduction pattern shown in Figure~\ref{fig:pattern}. We compare our approach with current state-of-the-art methods, which achieve acceleration by compressing the number of visual tokens, specifically FastV~\cite{FASTV} and VTW~\cite{VTW}.
As shown in Table~\ref{tab:main}, our approach achieves comparable or superior performance to these token compression methods while reducing floating point operations (FLOPs) by approximately 30\%. Additionally, our method and token compression methods address acceleration from different perspectives: token compression methods aim to reduce the number of visual tokens, whereas our approach focuses on lowering the computation required per visual token. This distinction indicates that the two methods are orthogonal and can be combined to achieve further acceleration. Table~\ref{tab:main} demonstrates this synergy: when applying our approach alongside FastV~\cite{FASTV} to reduce FLOPs by about 50\%, model performance significantly surpasses that of FastV~\cite{FASTV} alone with a higher compression rate across most benchmarks.

It is important to emphasize that our primary objective is to demonstrate the effectiveness of per-token computation reduction as an alternative acceleration approach, rather than to establish its superiority over token compression approaches.
In fact, each approach is suited to different scenarios, as shown in Table~\ref{tab:main}. In cases of high information density within images, especially in text-rich contexts like OCRBench~\cite{OCRBench}, the potential for visual token compression is limited, making per-token computation reduction more effective. Conversely, in scenarios with lower information density, such as MME~\cite{MME}, reducing the number of visual tokens offers a higher upper bound for acceleration. This complementarity between approaches suggests that the optimal choice of acceleration strategy should be context-dependent, with the potential for combined implementation in hybrid solutions.

\subsection{Ablation Studies}

\begin{figure}[t]
  \centering
  \includegraphics[width=1.0\linewidth]{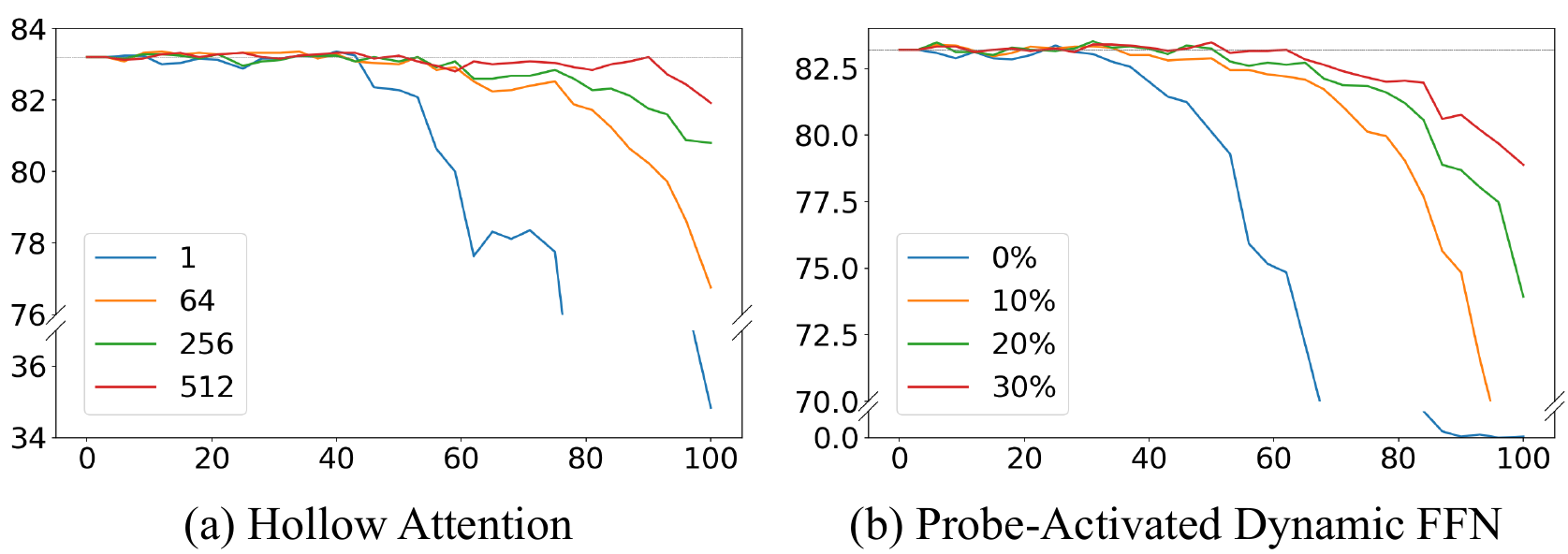}
  \caption{Ablation studies on (a) the attention range in Hollow Attention and (b) the proportion of activated parameters in Probe-Activated Dynamic FFN, evaluated on ChartQA by InternVL2-8B. The x-axis indicates the percentage of layers with reductions applied, while the y-axis reflects the model’s performance.}
  \label{fig:a3}
\end{figure}

\textbf{Ablation Studies on the Extent of Reduction in Self-Attention and FFN.}
To comprehensively evaluate the impact of reducing self-attention and FFN operations, we conduct ablation studies focusing on two factors: the proportion of activated parameters in Probe-Activated Dynamic FFN and the attention range in Hollow Attention, as illustrated in Figure~\ref{fig:a3}. The results show that as the extent of reduction decreases (i.e., as the proportion of activated parameters or the attention range increases), more layers can be reduced without significantly impacting model performance. These two hyperparameters are selected by trading off efficiency and effectiveness.

\textbf{Ablation Study of Layer Ranking Strategies.}

\begin{figure}[t]
  \centering
  \includegraphics[width=1.0\linewidth]{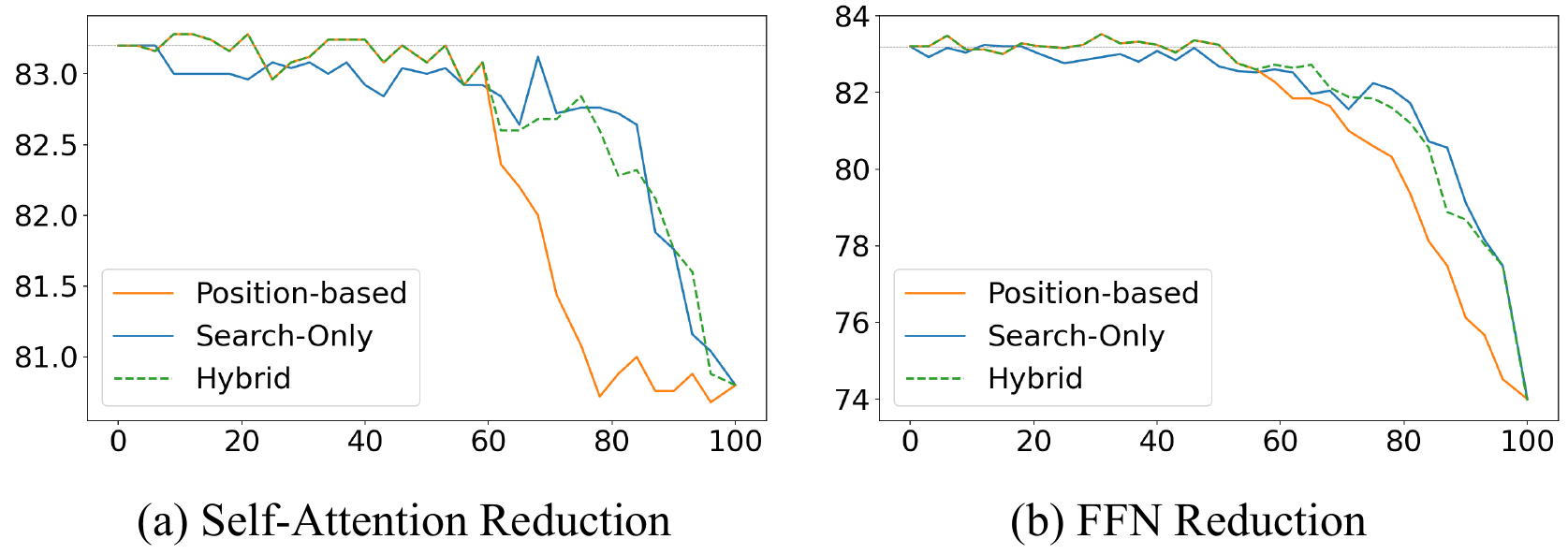}
  \caption{Performance comparison of different layer ranking strategies, evaluated on ChartQA by InternVL2-8B. The first half of the orange line (Position-based) overlaps with and is obscured by the green line (Hybrid).}
  \label{fig:a1}
\end{figure}

We compared three layer ranking strategies: (1) Position-based Strategy, which assigns the highest rank to the last layer and progressively decreases the rank toward the first layers; (2) Search-only strategy, which relies solely on Algorithm~\ref{alg:layer_ranking}, with no layers pre-assigned ranks; and (3) Hybrid strategy, where the last \(L_p\) layers are pre-assigned the highest rank, with the remaining layers ranked by Algorithm~\ref{alg:layer_ranking}. As shown in Figure~\ref{fig:a1}, when reduction is applied to only a few layers, the position-based strategy outperforms the search-only strategy, indicating that later layers tend to exhibit higher redundancy in visual token processing. Additionally, the limited size of our validation set may not fully capture the true behavior of the models. As the number of reduced layers increases, the search-only strategy begins to yield better results. Therefore, we adopt the hybrid strategy, which combines the position-based strategy with the search-only strategy, to achieve better performance and reduce the number of evaluations required by Algorithm~\ref{alg:layer_ranking}.

\section{Conclusion}
\label{sec:conclusion}

In this paper, we present a systematic investigation into the redundancy of visual token processing, which plays a crucial role in the trade-off between performance and efficiency in mainstream MLLM architectures. Through careful analysis of existing MLLMs, we propose a new framework consisting of two key components: computational reductions for visual tokens and a layer ranking algorithm. These reductions are applied across various layer proportions to evaluate their impact on MLLM performance. Extensive experiments reveal that current decoder-only MLLMs exhibit significant redundancy in visual token processing within certain layers. This structured and clustered redundancy can be effectively leveraged, providing valuable insights for future architectural design. Furthermore, this work opens new perspectives on training-free acceleration strategies for MLLMs, suggesting that future improvements in model efficiency might benefit from considering both token-level compression and computation-level optimization.

\section*{Limitations}

Determining the layer rank for reduction through search in the validation set presents two limitations. First, it requires constructing a validation set and performing hundreds of evaluations. Additionally, to reduce computational resource demands, we use a limited-scale validation set and a greedy search-based algorithm, which may fail to identify the optimal combination of layers for reduction. Therefore, improvements to the Layer Ranking Algorithm or exploration of alternative features for determining layer reduction priorities warrant further investigation.

\section*{Acknowledgements}

This research is supported in part by the National Natural Science Foundation of China (Grant No.: 62476093, 62441604).

\bibliography{main}

\appendix

\end{document}